\documentclass[journal]{IEEEtai}
\usepackage[figuresright]{rotating}
\usepackage[linesnumbered,ruled,vlined]{algorithm2e}
\usepackage{algorithmic}
\usepackage{graphicx}
\usepackage{url}
\usepackage{amsmath}
\usepackage{subfigure}
\usepackage{threeparttable}
\usepackage{bm}
\usepackage{float}
\usepackage{color}
\usepackage{booktabs}  
\usepackage{multirow}
\usepackage{setspace}
\usepackage{caption}
\usepackage{epstopdf}
\usepackage{booktabs}
\usepackage{threeparttable}
\usepackage{pifont}
\usepackage{makecell}
\usepackage{cite}
\usepackage{amsfonts}
\usepackage{longtable}
\usepackage{amsmath}

\begin{document}

\title{Learn Once Plan Arbitrarily (LOPA): Attention-Enhanced Deep Reinforcement Learning Method for Global Path Planning}
%
%


\author{
	\IEEEauthorblockN{Guoming Huang$^{1,2}$, Mingxin Hou$^2$, Xiaofang Yuan$^2$, Shuqiao Huang$^{1}$, Yaonan Wang$^2$}
	
	\thanks{Mingxin Hou and Guoming Huang contributed to the work equally and should be regarded as co-first authors. Prof. Xiaofang Yuan is the corresponding author. This work is done in the Autonomous Driving Lab hold by Prof. Xiaofang Yuan at Hunan University.}
	
	\thanks{$^1$ The author is with the School of Electronic Engineering and Automation, Guilin University of Electronic Technology, Guilin 541004, China. Shuqiao Huang's e-mail: 21082304036@mails.guet.edu.cn.}
	
	\thanks{$^2$ The authors are with the College of Electrical and Information Engineering, Hunan University, Changsha 410082, China. E-mail: \{huangguoming, mx\_hou, yuanxiaofang, Yaonan\}@hnu.edu.cn).}

}


\maketitle

\begin{abstract}

Deep reinforcement learning (DRL) methods have recently shown promise in path planning tasks. However, when dealing with global planning tasks, these methods face serious challenges such as poor convergence and generalization. To this end, we propose an attention-enhanced DRL method called LOPA (Learn Once Plan Arbitrarily) in this paper. Firstly, we analyze the reasons of these problems from the perspective of DRL's observation, revealing that the traditional design causes DRL to be interfered by irrelevant map information. Secondly, we develop the LOPA which utilizes a novel attention-enhanced mechanism to attain an improved attention capability towards the key information of the observation. Such a mechanism is realized by two steps: (1) an attention model is built to transform the DRL's observation into two dynamic views: local and global, significantly guiding the LOPA to focus on the key information on the given maps; (2) a dual-channel network is constructed to process these two views and integrate them to attain an improved reasoning capability. The LOPA is validated via multi-objective global path planning experiments. The result suggests the LOPA has improved convergence and generalization performance as well as great path planning efficiency.
\end{abstract}

\begin{IEEEImpStatement}
	
When applying DRL to address the global path planning problem, performance often tends to be sub-optimal. This is primarily attributed to the dynamic nature of terrain alterations, resulting in a boundless state space. Consequently, when the DRL agent is confronted with less familiar states, poor action in decision-making frequently arises (that leads to poor convergence and generalization). To solve this fundamental problem, we propose an approach named LOPA. This approach leverages an attention-enhanced mechanism to bolster the network's capacity for terrain comprehension, thereby leading to notable enhancements in the efficacy of global path planning outcomes. Experiments show that LOPA exhibits accelerated convergence and superior planning efficacy compared to traditional DRL method. Simultaneously, LOPA outperforms traditional path planning approaches (A*, RRT) in terms of efficiency. In addition, the proposed method can be transferred to other reinforcement learning scenarios to deal with the problem of infinite state space.
	
\end{IEEEImpStatement}


%
\IEEEpeerreviewmaketitle

\section{Introduction}

\subsection{Motivation}

\IEEEPARstart{I}{n} recent years, deep reinforcement learning (DRL) has emerged as a promising solution for solving NP-hard problems \cite{ref6Solving, ref3OpenAI, ref5NI2021324}, attracting considerable attention from DRL researchers \cite{ref7Miki8659266}. One such problem is path planning, which also falls under the category of NP-hard problems. To date, several DRL-based path planning methods have been published \cite{8GUO2021479,20Zhao9533570,24Gao20195493,14Hu9244647}, which have shown significant results in different scenarios. However, little attention has been given to global path planning. In such cases, DRL is hard to grab the key information in global maps, thereby fail to learn a comprehensive planning policy, encountering serious convergence and generalization problems.


To overcome this challenge, we propose an attention-enhanced DRL method called LOPA (Learn Once Plan Arbitrarily) in this work. LOPA employs a novel attention-enhanced mechanism to attain an improved comprehension capability to the key terrain information, hence can devise high-quality global path planning strategies.


\subsection{Related Works}

Path planning has been extensively studied with traditional methods \cite{40Ravankar9286439,49Wang9727523,33Xu2020,34Guo2020GlobalPP,51Norouzi6385821}. Although the A* algorithm and Dijkstra method can find the optimal path, they require searching a relatively large area \cite{37Ammar,51Norouzi6385821}. The Ant Colony method is prone to becoming stuck in local minima \cite{54Ma8540402}, while the Rapidly-Exploring Random Trees (RRT) method exhibits unstable planning performance \cite{40Ravankar9286439,49Wang9727523}. In summary, traditional methods either search a large area to obtain a better global solution or narrow the search area to achieve a sub-optimal solution. They struggle to strike a balance between search time and solution quality \cite{H3DM9408393}.

In recent years, DRL theory has demonstrated remarkable potential for solving path planning problems \cite{ref6Solving}. Numerous DRL methods have been proposed \cite{8GUO2021479,11Jiang8853432,9Cimurs,18Wang8600371,25Jiang9536098,27Zeng9269419}, which have achieved significant progress in various tasks such as 3D path planning for unmanned aerial vehicles \cite{8GUO2021479} and navigation path planning for ships \cite{3Guo20020426}. Among these methods, the primary difference lies in the design of input (observation) used by the DRL method rather than the specific DRL algorithm utilized. Therefore, these methods can be broadly categorized into three groups based on their observation strategy: (1) orientation information only; (2) local map and orientation information; and (3) entire map with a moving object. In the following, the related work will be introduced from these three aspects.

\subsubsection{Path planning using the orientation information}

The orientation information utilized by these methods typically involves a description of the current point with respect to obstacles or the objective, primarily considering path traversability and safety \cite{8GUO2021479,20Zhao9533570,11Jiang8853432,3Guo20020426}. For instance, Tong et al. \cite{8GUO2021479} propose an ultrasonic range finder-based DRL method to identify a collision-free and efficient path for unmanned aerial vehicles in challenging environments. Similarly, Guo et al. \cite{3Guo20020426} develop a DRL approach that utilizes the heading angle, distance between the ship and obstacles, and deviation angle between the ship and target to plan navigation paths. Focusing solely on traversability without considering globality can result in planning with excessive distance \cite{8GUO2021479}. In practical application, the optimal path should satisfy both the requirement of traversability and minimize the total distance traveled.

\subsubsection{Path planning using the local map and orientation information}

Compared to the previous group, the methods in this category integrate local and orientation information to construct the observation, which can improve the convergence speed of DRL algorithms \cite{13DBLP,10Xie9348925,9Cimurs,18Wang8600371}. For instance, a DRL method for city-wide navigation tasks is proposed in \cite{13DBLP}, where despite the large planning area, only local observations are fed into the network. However, this approach only prioritizes traversability rather than the total distance traveled. Similarly, \cite{18Wang8600371} presents an online DRL-based path planning algorithm that enables autonomous navigation in complex virtual environments using local observations. While these methods achieve improved convergence speeds, they still plan paths without considering globality, resulting in average path quality \cite{10Xie9348925}.

\subsubsection{Path planning using the entire map with a moving object}

Considering only traversability is insufficient for achieving optimal path planning. Therefore, researchers have attempted to enhance path globality by leveraging global information in DRL-based approaches \cite{24Gao20195493,14Hu9244647,25Jiang9536098,27Zeng9269419}. For example, Gao et al. \cite{24Gao20195493} propose a method that utilizes a global view as the observation to enable path planning in different indoor scenes. Hu et al. \cite{14Hu9244647} investigate collision-free path planning for multiple robots by partitioning the exploration area to reduce repeated exploration. By incorporating global information, these methods can find the shortest paths. However, they are primarily designed to solve maze navigation problems in small-sized indoor environments (e.g., 20$\times$20). For larger maps (e.g., 50$\times$50 or bigger size), the existing DRL methods still face significant convergence and generalization issues \cite{24Gao20195493}.

In summary, the research of DRL methods in local path planning is relatively mature. However, the research on global path planning is insufficient. When faced with the global path planning task in slightly larger maps, DRL agents need to know the necessary global information. However, the huge observation space can cause problems of convergence and generalization. These have become an important fundamental challenge for DRL methods to be widely used in global path planning, and needs to be solved urgently.


\subsection{Contributions}
To this end, we carry out this work to build an improved DRL method for global path planning tasks. The contributions of this work are listed as follows.

\begin{itemize}
	
	\item{We focus on the fundamental problems (convergence and generalization) faced by DRL method when it is used for global path planning. The research results can provide technical reference for the research community.}

	\item{We propose the LOPA which introduces an attention-enhanced mechanism. In the LOPA, an attention model and a dual-channel network are constructed to attain an improved attention capability.}
	
	\item{We validate the LOPA via multi-objective 2.5D global path planning experiments which are a popular path planning topic \cite{55Sutoh7059362, 56Azizi7731923, 57Ji8466595}. Through such experiments, it suggests that the proposed method can better solve the convergence and generalization problem.}

\end{itemize}

\subsection{Organization of this paper}
The remainder of this paper is arranged as: Section \uppercase\expandafter{\romannumeral2} is the preliminaries including the construction of planning environment and the problem analysis; Section \uppercase\expandafter{\romannumeral3} introduces the design of the LOPA method; Section \uppercase\expandafter{\romannumeral4} provides the experiments;  At last, the conclusion and prospect are given in Section \uppercase\expandafter{\romannumeral5}.

\section{Preliminaries}

\subsection{2.5D path planning environment}
The proposed method is intended to be validated through experiments in 2.5D multi-objective global path planning. To this end, we have constructed a 2.5D path planning simulation environment using the GYM platform. In this environment, the path planning task is designed as a game in which Mario searches for mushrooms, with the mushroom representing the target location and Mario representing the DRL agent. The map used in the simulation is a randomly generated 2.5D map, where the X-axis represents latitude, the Y-axis represents longitude, and the Z-axis represents altitude. When given a path, the simulation environment is able to return the distance and energy consumption required for the path. Meanwhile, the DRL agent have eight direction to select during the planning process.

\subsection{Problem analysis}

In this section, the causes of convergence and generalization problems will be analyzed from the observational perspective. 

Regardless of whether value-based methods or policy gradient methods are used to implement path planning, both fundamentally require learning a network to estimate the cost of planning actions based on provided states. Taking Q-learning as an example, the learning process can be understood as constructing a mapping from states to Q-values. To establish this relationship, a mapping function based on deep neural networks is formulated as follows:
\begin{equation}
	Q = Net(\mathbf{S}, \mathbf{\theta})
\end{equation}
where \(Net(\cdot)\) denotes the mapping function based on deep neural networks, and \(\mathbf{\theta}\) represents the network parameters. \(\mathbf{S}\) is the given state which generally consists of map \(\mathbf{M}\), current position \(X_c\), and destination \(X_d\). \(\mathbf{M}=\begin{bmatrix} x_{1,1}& \cdots &x_{m,1}\\ \vdots & \ddots & \vdots \\ x_{1,n}& \cdots & x_{m,n} \end{bmatrix}\), where \(m\) is the number of rows in the digital elevation map and \(n\) is the number of columns. Here, Eq. (6) can be abstractly rewritten as:
\begin{equation}
	Q = \mathbf{\Gamma}(\mathbf{W \cdot S})
\end{equation}
where \(\mathbf{\Gamma}(\cdot)\) is a complex mapping function, \(\mathbf{W}\) is the weight matrix. Furthermore, Eq. (7) can be expanded as:
\begin{equation}
	Q = \mathbf{\Gamma}(\mathbf{w_1 M} + \mathbf{w_2 X_c} + \mathbf{w_3 X_d})
\end{equation}
where \(\mathbf{w_i}\) represents the corresponding weights. 

Here, we discuss two kinds of scenarios, (1) planning on a single map, and (2) planning on random maps. 

\subsubsection{Planning on a single map}
Without loss of generality, to determine \(\mathbf{\theta_i}\) can be regarded as solving a homogeneous system of equations: 
\begin{equation}
	\begin{split}
		\begin{cases}
			Q_1 = \mathbf{\Gamma}(\mathbf{w_1 M^1} + \mathbf{w_2 X_c^1} + \mathbf{w_3 X_d^1}) \\
			Q_2 = \mathbf{\Gamma}(\mathbf{w_1 M^2} + \mathbf{w_2 X_c^2} + \mathbf{w_3 X_d^2}) \\
			Q_3 = \mathbf{\Gamma}(\mathbf{w_1 M^3} + \mathbf{w_2 X_c^3} + \mathbf{w_3 X_d^3}) \\
			\quad \qquad \vdots \\
			Q_k = \mathbf{\Gamma_k}(\mathbf{w_1 M^k} + \mathbf{w_2 X_c^k} + \mathbf{w_3 X_d^k} )
		\end{cases}
	\end{split}
\end{equation}
Due to \(\mathbf{M^1 = M^2 = M^3 = \cdots = M^k}\), the column vectors containing map information remain unchanged, so that Eq. (3) can be simplified as:
\begin{equation}
		Q = \mathbf{\Gamma}(\mathbf{w_2 X_c} + \mathbf{w_3 X_d}) + C
\end{equation}
where \(C\) is a constant. As indicated in Eq. (5), map information does not participate in the calculation of \(Q\), which leads to the DRL agent ultimately being unable to understand the terrain on maps. 

\begin{figure}[htp]
	\centering
	\includegraphics[width=0.98\columnwidth]{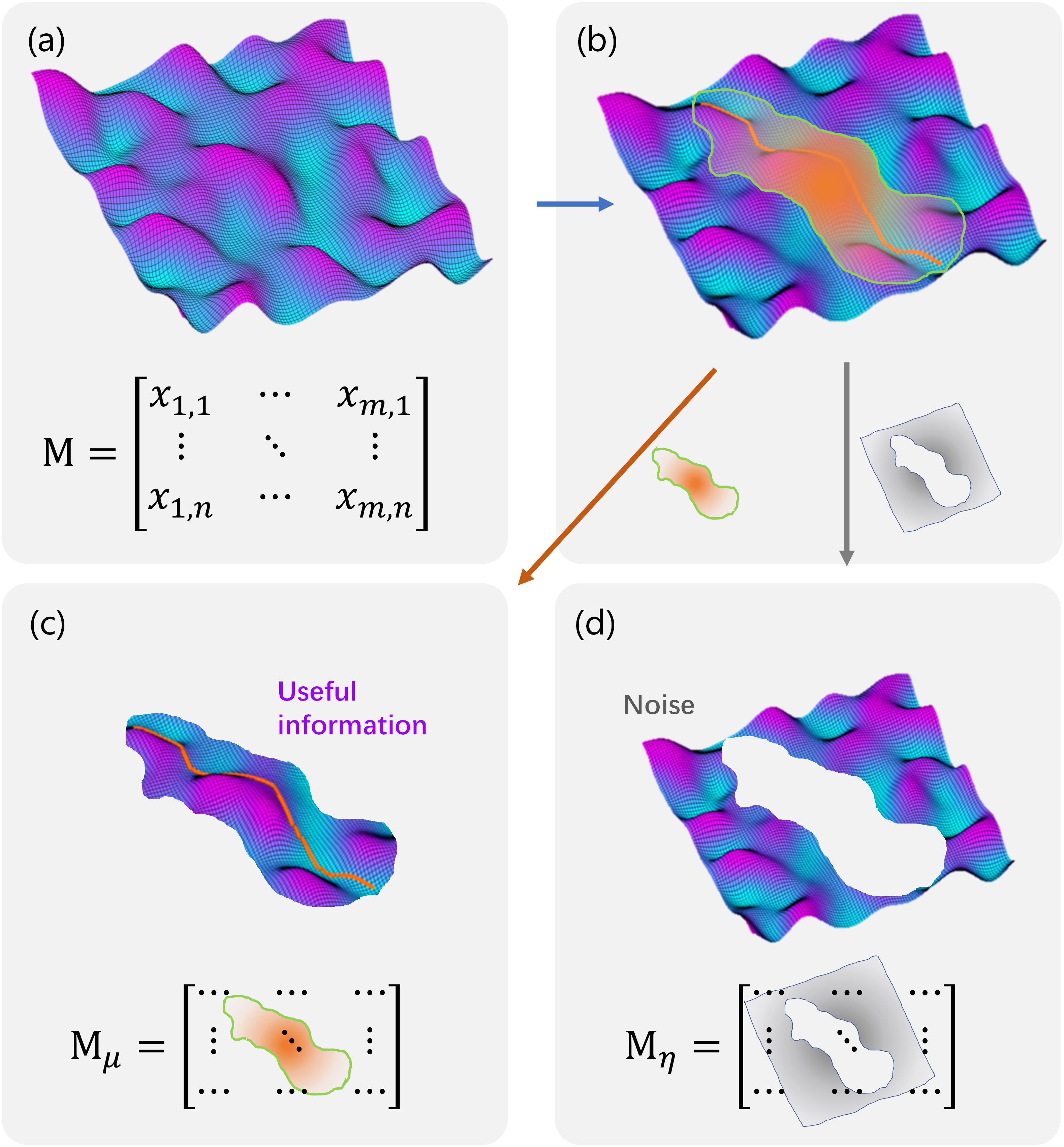}
	\caption{The utilization of map information in planning. (a) The given map. (b) The planning process. (c) The useful information. (d) The noise. }
\end{figure}

\begin{figure*}[htp]
	\centering
	\includegraphics[width=1.9\columnwidth]{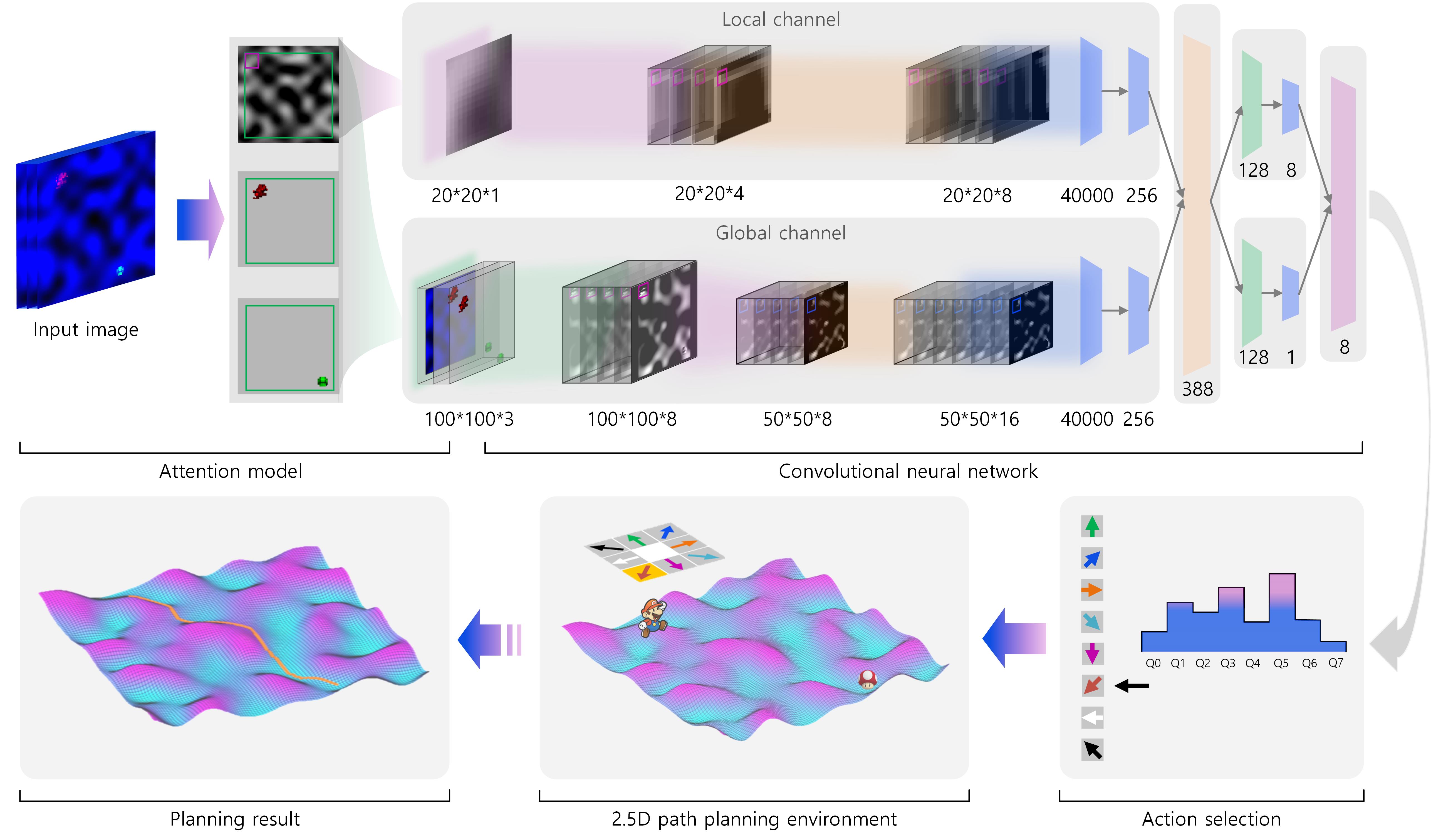}
	\caption{The LOPA method for 2.5D path planning. The upper row illustrates the framework of LOPA. The lower row depicts the path planning process using LOPA. Starting from the right side, LOPA selects an action, which is then executed by the environment, yielding a new state. After several iterations, the optimal path is obtained through repeated action selection and execution.}
\end{figure*}

\subsubsection{Planning on random maps}
Though the map is utilized in calculating the \(Q\), but the network is still difficult to converge and generalize. The issue lies in the determination of \(\mathbf{w_1}\), where it is apparent that not all the terrain information in \(\mathbf{M}\) contribute to the calculation of \(Q\). Regardless of the A* search \cite{51Norouzi6385821} or intuitive human experience, obtaining optimal paths does not require using the entire map in reasoning. As shown in Fig. 1, the information mainly used for reasoning is concentrated between the current point and the destination \cite{51Norouzi6385821} (see sub-figure (c)), while most of the other terrains are noise (see sub-figure (d)). Accordingly, Eq. (8) can be re-whiten as:
\begin{equation}
	Q = \mathbf{\Gamma}(\mathbf{w_{\mu} M_{\mu}} + \mathbf{w_{\eta} M_{\eta}} + \mathbf{w_2 X_c} + \mathbf{w_3 X_d})  
\end{equation}
Here, assume \(\mathbf{M} = \mathbf{M_{\mu}} + \mathbf{M_{\eta}}\), \(\mathbf{w_{\mu}}\) and \(\mathbf{w_{\eta}}\) are the corresponding weights split from \(\mathbf{w_1}\). \( \mathbf{M_{\eta}}\) is regarded as the noise, making the network hard to converge. Otherwise, if the network converges, the learned \(\mathbf{w_{1\eta}}\) may cause an over-fitting issue, leading to poor generalization. Based on the above analysis, it is necessary to find a way that can reduce the interference of noise (irrelevant information) on the map and improve the utilization of the key information.


\section{Design of LOPA}
In this section, the structure of LOPA, modeling of reward function and training strategy are introduced, respectively.

\subsection{Structure of LOPA}

The LOPA is developed by utilizing an attention-enhanced mechanism. As depicted in Fig. 2, the LOPA consists of an attention model and a dual-channel network. The attention model transforms the input into two dynamic views, namely the local view and global view. The dual-channel network firstly processes these views separately and combine them to facilitate effective policy-reasoning.

\subsubsection{Attention model}

According to the previous analysis, if we can remove \(\mathbf{M_{\eta}}\) from \(\mathbf{M}\) and only keep \(\mathbf{M_{\mu}}\) as the observation, then the mentioned problems would be solved. That is to say, we need to improve the attention capability of DRL towards \(\mathbf{M_{\mu}}\), so that it ignores the irrelevant terrain information \(\mathbf{M_{\eta}}\). However, the shape of \(\mathbf{M_{\mu}}\) is obviously related to the specific given map, which is hard to be determined. Here, we propose a simple but effective solution, which is to define \(\mathbf{M_{\mu}}\) as a dynamic rectangular window. This effectively ensures that the key information is used and removes most of the useless information. From the perspective of neural network training, we compress the solution space and reduce the learning difficulty of the network. In practice,  \(\mathbf{M}\) is modified as \(\begin{bmatrix} 0 & \cdots & 0 \\ \vdots & \mathbf{M_{\mu}} & \vdots \\ 0 & \cdots & 0 \end{bmatrix}\). \(\mathbf{M_{\mu}}\) is actually the mentioned global dynamic view. Meanwhile, the local view is built to further improve the attention of DRL to the key terrain nearby the current position (see Fig. 3). The specific operation to the global view and the local view is introduced as follows.

\begin{figure*}[htp]
	\centering
	\includegraphics[width=1.8\columnwidth]{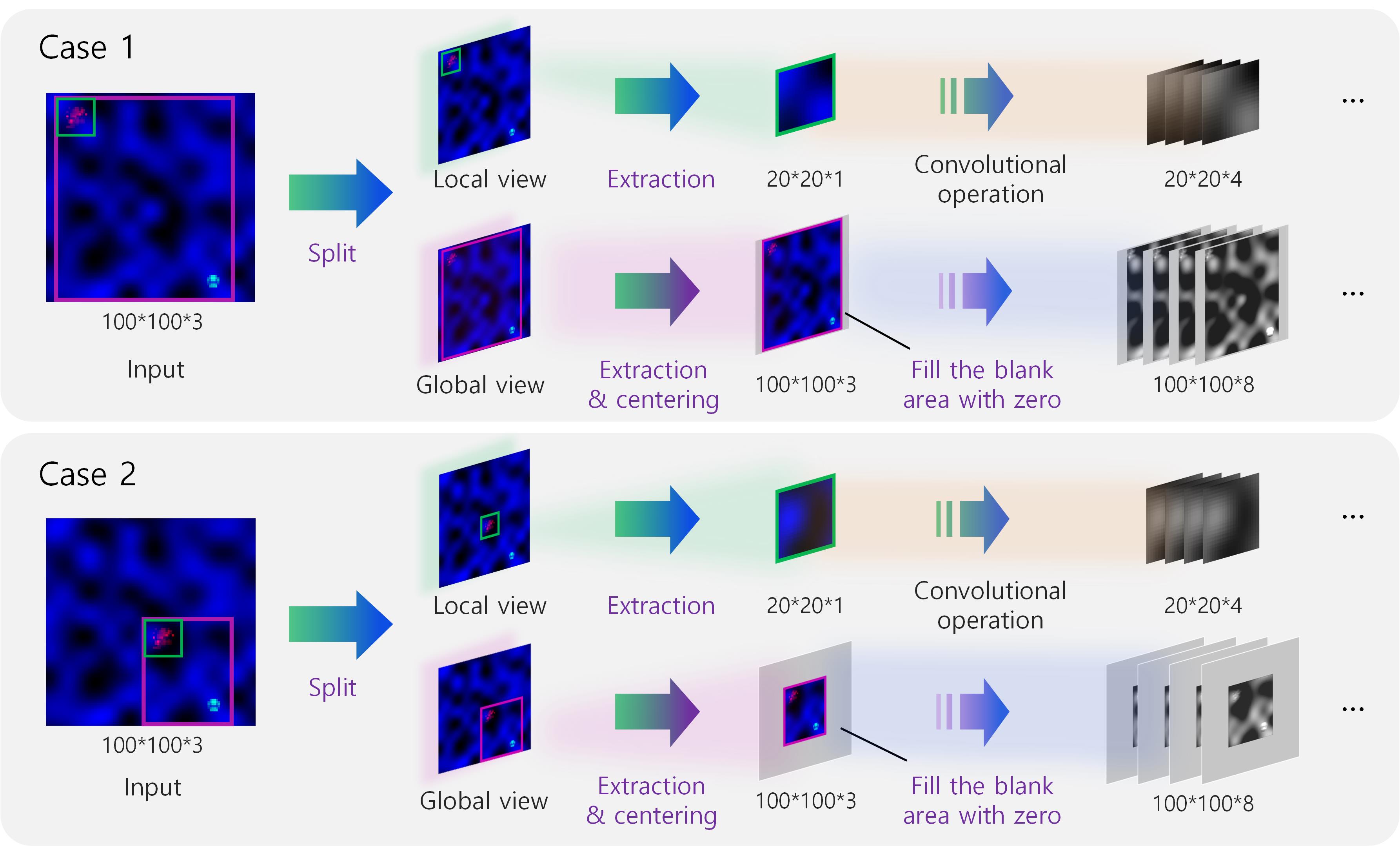}
	\caption{Operation cases of attention model. Case 1 depicts Mario situated at the outset, whereas Case 2 shows Mario's movement to a specific location on the map. These scenarios indicate that by leveraging the attention model, the LOPA can consistently concentrate on crucial regions during the planning phase, thereby attaining superior convergence and planning efficacy.}
\end{figure*}

\begin{itemize}
	
	\item{Regarding the global view, we employ a three-step process to convert the original input into a new "dynamic" representation: (1) Create a rectangular field that encompasses a diagonal line connecting the current position and the target. (2) Extend the border of the rectangular field by 10 steps. (3) Center the rectangular field and fill up the other regions with 0. In practice, we maintain the size of the global view at 100*100*3. The terrain, Mario, and mushroom are placed on distinct layers (see Fig. 2).}
		
	\item{Regarding the local view, we extract a small region surrounding the Agent's current location. This area is updated at every planning step, facilitating the LOPA's intricate examination of nearby terrain. In practice, we center the current position, expand symmetrically around the map to form a 20*20*1 rectangular field.}
	
\end{itemize}

\subsubsection{Dual-channel network}

To achieve enhanced policy-reasoning capabilities, the dual-channel network is designed to work in tandem with the attention model. The dual-channel network takes the local and global views provided by the attention model as input and processes this information separately using distinct channels via convolutional operations. The resulting convolutional outputs are then concatenated to make decisions.

The structure of the dual-channel network is illustrated in Fig. 1 and comprises three components: the input layer, convolutional backbone, and fully connected layers. Alternatively, we can view the dual-channel network as comprising two distinct channels: a local channel and a global channel. The local channel takes the local view as input, sized at 20*20*1, while the global channel accepts the global view as input, sized at 100*100*3.


The convolutional backbone of the global channel features two convolutional layers: the first convolutional kernel measures 8*(3*3*3), while the second measures 16*(3*3*8). Both layers utilize ReLu activation functions, and a maximum pooling operation is applied between the two layers. The pooling size is set at 2*2, and the step is 3. The convolutional backbone of the local input channel comprises two convolutional layers as well: the first convolutional kernel is sized as 4*(3*3*1), whereas the second is sized as 10*(3*3*4). Similarly, both layers utilize ReLu activation functions. Given that the local channel aims to provide terrain-specific details, no pooling layers are used in this case.

The fully connected layers consist of three parts: The first connects the final layers of the two channels and then diverges into two branches (see Fig. 1). Each branch consists of two layers \cite{dueling}. One branch features 128 nodes in its first layer and 8 nodes in its second layer, while the other has 128 nodes in its first layer and 1 node in its second layer. The activation functions used in the first and second layers are ReLu and linear, respectively. Finally, the two branches converge at the last layer, which comprises 8 nodes. The output of the last layer represents the Q values of actions.

\subsection{Reward function and Hybrid exploration strategy}

\subsubsection{Reward function}
In this work, we verify the performance of LOPA through multi-objective 2.5D global path planning experiments in which the distance and energy-consumption are both considered \cite{34Guo2020GlobalPP, 35DAVOODI2015568}. Therefore, specialized reward functions need to be designed. In the context of multi-objective path planning, it is crucial to consider not only terrain traversability but also globality, achieving a favorable balance between energy consumption and distance traveled.

\subsubsection{Hybrid exploration strategy}

The exploration strategy is comprised of both heuristic knowledge and random exploration tactics. The heuristic knowledge provides valuable direction information, which serves to encourage the Agent's movement towards its intended target. During the training process, the probability of selecting the heuristic action commences at a relatively high rate (0.5) and gradually diminishes to 0.01 towards completion. Concurrently, the probability of the random action conforms to the same pattern as the heuristic action. In contrast, the probability of the LOPA-suggested action increasingly rises from 0 to 0.98. Besides, we also adopt the prioritized experience replay \cite{PER} method to train the LOPA.

\section{EXPERIMENT}

In this section, experiments have been conducted to demonstrate the advantages of LOPA from three perspectives, as presented in TABLE I: (1) Convergence performance validation of LOPA; (2) 
Generalization performance validation of LOPA in comparison with a local-map-based DQN (noted as LDQN), which takes the local view, current position and the destination as input; and (3) Extra generalization performance validation of LOPA in comparison with traditional methods.

%
%

\begin{table}[htp]
	\fontsize{7.5}{8}\selectfont
	\newcommand{\tabincell}[2]{\begin{tabular}{@{}#1@{}}#2\end{tabular}}   
	\centering  
	\renewcommand\arraystretch{1.8}
	\caption{Experiment Setting}
	\begin{tabular}{c|c|c}
		\Xhline{1.2pt} 
		\textbf{Experiment} & \textbf{Goal} & \textbf{Figures and Tables} \\  
		\hline
		1 &  \tabincell{c}{Compare the LOPA with \\ traditional dueling-DQN to \\ verify that the LOPA can \\ converge stably and fast} & Fig.4 \\
		
		\hline
		2 & \tabincell{c}{Compare the LOPA with \\ LDQN from the globality \\perspective} & Fig.5 - Fig.8  \\
		
		\hline
		3 & \tabincell{c}{Compare the LOPA with \\ the A* method \cite{37Ammar}, \\ the H3DM method \cite{H3DM9408393}, and \\ the improved RRT method \cite{40Ravankar9286439}} & \tabincell{c}{Fig.9 - Fig.12 and \\ TABLE III - TABLE VI} \\
		\Xhline{1.2pt}		
	\end{tabular}
\end{table}

\subsection{Experiment Setting}

All 2.5D maps are randomly generated, with dimensions ranging from 0 to 100 for length and width, and randomly ranging from 0 to 5 for height. These maps can be considered as zoomed in views of a 10 km x 10 km area with an altitude of 200 m. The experiments were run on a server with NVIDIA GTX 3090 graphics card, and Intel Core i9-10980XE CPU; software was used in Anaconda 64-bit.

Experiment 1 involves training the network on a single map with multiple tasks. Multi-task refers to randomly setting a set of starting points and targets during training. All other experiments use a multi-map and multi-task training mode, where each training episode randomly samples a map from a set of 50 maps, and sets the starting point and target locations randomly. Additionally, the unit of energy is u, which is proportional to J \cite{H3DM9408393}. The hyper-parameters used in these experiments are listed in TABLE II.
\begin{table}[htp]
	\fontsize{7.5}{8}\selectfont   
	\centering  
	\renewcommand\arraystretch{1.8}
	\caption{Hyper-parameters}
	\begin{tabular}{cc}
		\toprule[1.2pt] 
		\textbf{Parameter} & \textbf{Value} \\  
		\midrule
		Learning rate & 0.0005 \\
		\cline{1-2}
		Discount factor & 0.99\\
		\cline{1-2}
		Replay memory   &  50000 \\
		\cline{1-2}
		Batch size   &  128 \\
		\cline{1-2}
		Maximum episodes &  50000 \\
		\cline{1-2}
		Batch number per training & 20\\
		\cline{1-2}
		Batch size of target experiences    &  10 \\
		\cline{1-2}
		Initial probability of random action & 0.4 \\
		\cline{1-2}
		Initial probability of heuristic action  & 0.4 \\
		\cline{1-2}
		Initial probability of DQN action  & 0.2 \\
		\cline{1-2}
		Final probability of DQN action  & 0.99 \\
		
		\bottomrule[1.2pt]  
	\end{tabular}
\end{table}

\begin{figure*}[htp]
	\centering
	\includegraphics[width=1.9\columnwidth]{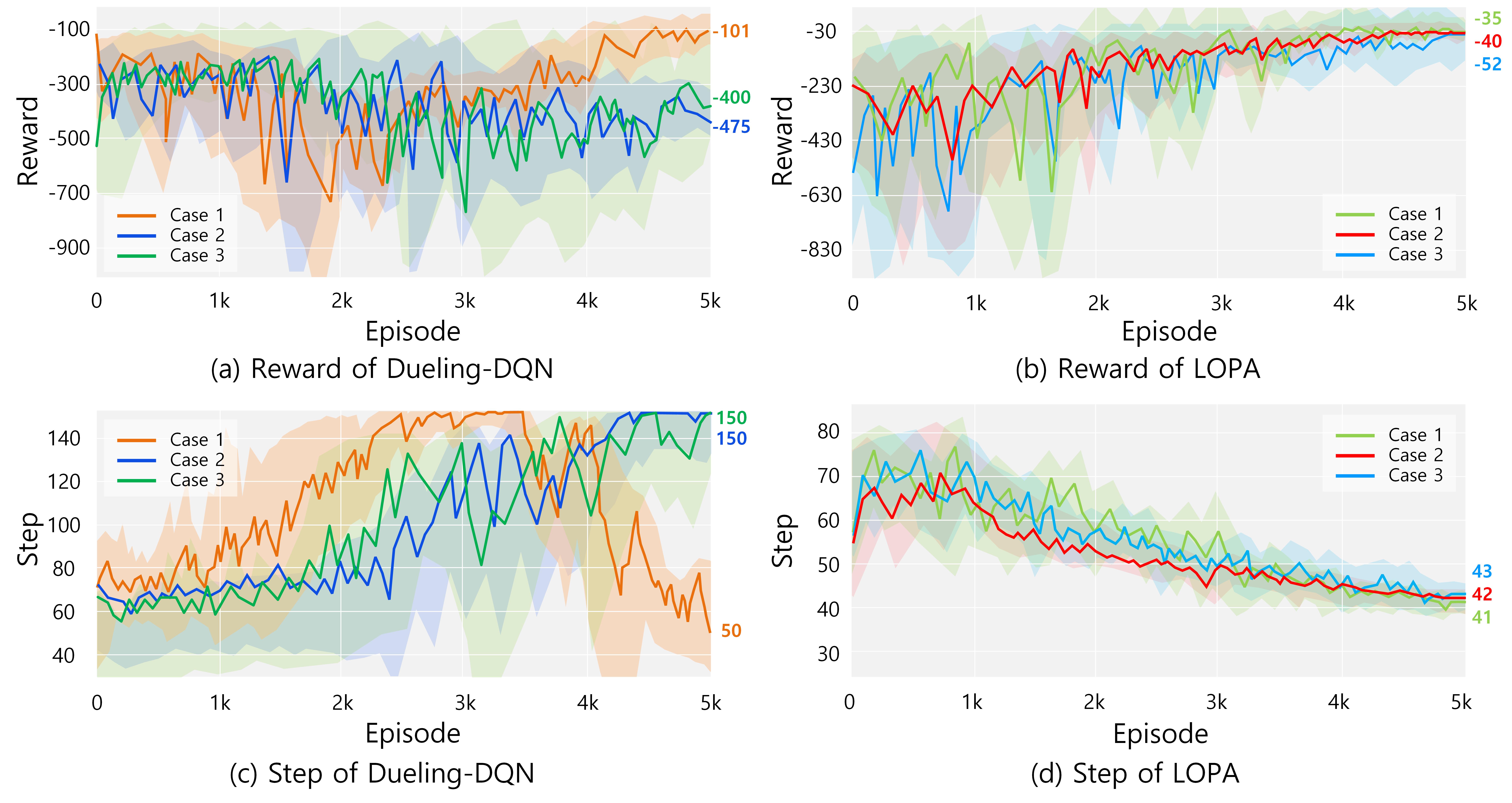}
	\caption{Performance curves of different methods during training on a 50*50 map. At the left side the convergence performance of the dueling-DQN is in chaos and unstable, whereas at the right side the LOPA shows significantly stable convergence performance.}
\end{figure*}

\subsection{Experiment 1}

\subsubsection{Setting}

In this experiment, the LOPA and Dueling-DQN are trained for multitasking. It should be noted that the Dueling-DQN struggles to converge when the map size is 100*100 or bigger, hence we chose a size of 50*50 for comparison here. The results of three training groups are depicted in Fig. 4. Each training consisted of 5k episodes with a maximum exploration step of 150 per episode.


\subsubsection{Result}

Fig. 4 demonstrates that the LOPA exhibits superior convergence performance compared to the dueling-DQN. Notably, the reward curves of LOPA demonstrate a remarkable capacity to converge from as early as 3,000 episodes, maintaining a steady level of convergence thereafter. In contrast, the reward curves of dueling-DQN in Cases 2 and 3 display instability, failing to achieve overall convergence. Case 1 shows some tendency towards convergence at approximately 4,500 episodes, albeit in a sub-optimal local minimum of -101. The step curves illustrate a substantial difference between the two approaches, with dueling-DQN requiring considerably more steps than LOPA. The number of failed episodes (i.e., episodes where the planning method fails to reach the target within 150 steps) for dueling-DQN gradually increases after 2,500 episodes, with almost all explorations in Case 2 experiencing failure after 4,200 episodes. Conversely, the curve of LOPA displays stable convergence, with a markedly higher success rate compared to dueling-DQN.

\subsubsection{Discussion}

Despite the dueling-DQN's ability to achieve convergence in 50*50 maps, the outcomes are noticeably unstable and unsatisfactory, with rewards post-convergence dramatically lower than those attained by the LOPA. As per the reward function, global planning knowledge acquisition is essential for maximizing total reward. In Case 1, the dueling-DQN exhibits convergence; however, the reward achieved -101, pales in comparison to that of LOPA (-35). This indicates that the dueling-DQN merely learns target attainment techniques rather than developing a comprehensive global perspective on planning.

Notably, the dueling-DQN functions on a static map input, with minimal changes in the current point during planning. The outcomes suggest that the dueling-DQN is potentially limited to metaphysical recollection of planning techniques at various points, rather than demonstrating comprehensive comprehension of the terrain. This accounts for the method's failure on larger maps (e.g., 50*50). Conversely, LOPA achieves stable and rapid convergence, highlighting the effectiveness of the attention model and dual-channel network in enhancing LOPA's reasoning capacities.


In addition, we have conducted extra convergence tests. For example, we trained the dueling-DQN on random maps of size 50*50, as well as single and random maps of size 100*100. In these cases, the dueling-DQN fails to converge while the LOPA converges stably. However, the comparison results in the single map scenario of size 50*50 are sufficient to demonstrate that LOPA's convergence performance is better than that of dueling-DQN. It should be noted that this part of the experiment focuses on DRL methods for global planning purposes. In fact, if the goal of global planning is not considered and only local maps are used as input, the current DRL methods do not suffer from convergence problems.

\subsection{Experiment 2}

This experiment aims to validate LOPA's generalization performance in comparison with the LDQN.

\subsubsection{Setting}

LDQN's input incorporates a local view spanning 20*20, the 2D distance between the current point and target, and the target orientation. Testing was performed using random maps obtained from the test set; four cases of results are presented in Fig. 5 through Fig. 8. Specifically, sub-figure (a) depict planned paths on 3D perspectives, sub-figure (b) display planned paths on 2D perspectives, sub-figures (c) and (d) showcase energy consumption and path length curves, respectively.

\subsubsection{Result}
The results show that the paths planned by the LOPA has better globality than the those of LDQN. The path of LDQN can reach the target, but sometimes directly crosses the peaks and valleys, and the path of LOPA avoids a large slope. In Fig. 5 (Case 1), near the target, the green path reaches target from the halfway up the peak, while the orange path traverses the very bottom of a valley at first and then climbs to reach the target. Fig.5(d) also shows that the energy consumption of the path planned by the LDQN increases dramatically by 1000 in comparison with the path of LOPA, due to overturning from the valley to the peak. In Fig. 7 (Case 3), at the larger peak between the start and target, the green path avoids the peak from the left to reach the target; The orange path goes around further to avoid the peak before reaching the target. In comparison with the green path, the orange path is 1000 further than the former. In summary, the globality of the paths planned by the LOPA method is better than that of the LDQN.

\begin{figure}[htp]
	\centering
	\includegraphics[width=0.98\columnwidth]{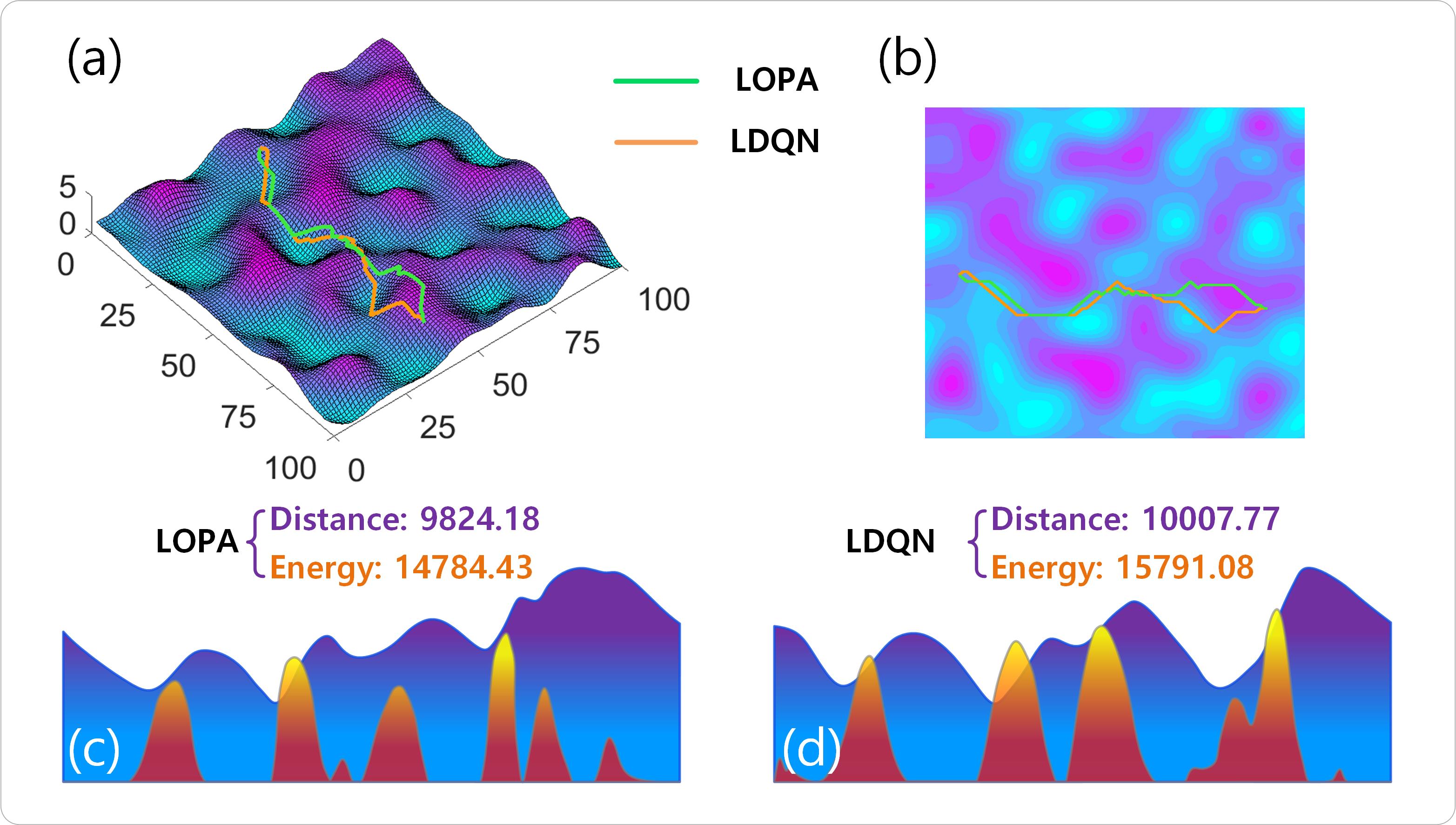}
	\caption{Case 1 of experiment 2}
\end{figure}

\begin{figure}[htp]
	\centering
	\includegraphics[width=0.98\columnwidth]{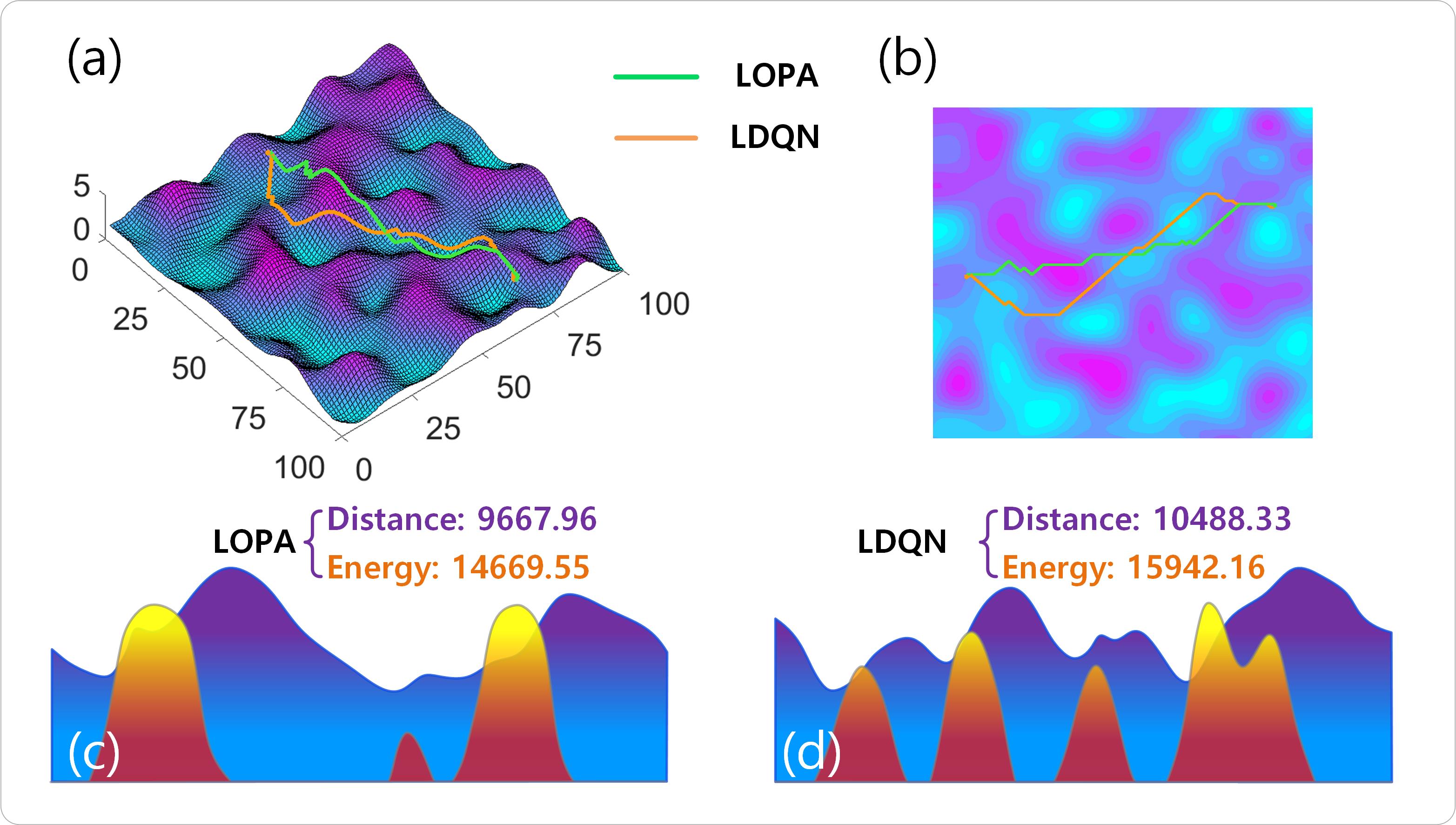}
	\caption{Case 2 of experiment 2}
\end{figure}

\begin{figure}[htp]
	\centering
	\includegraphics[width=0.98\columnwidth]{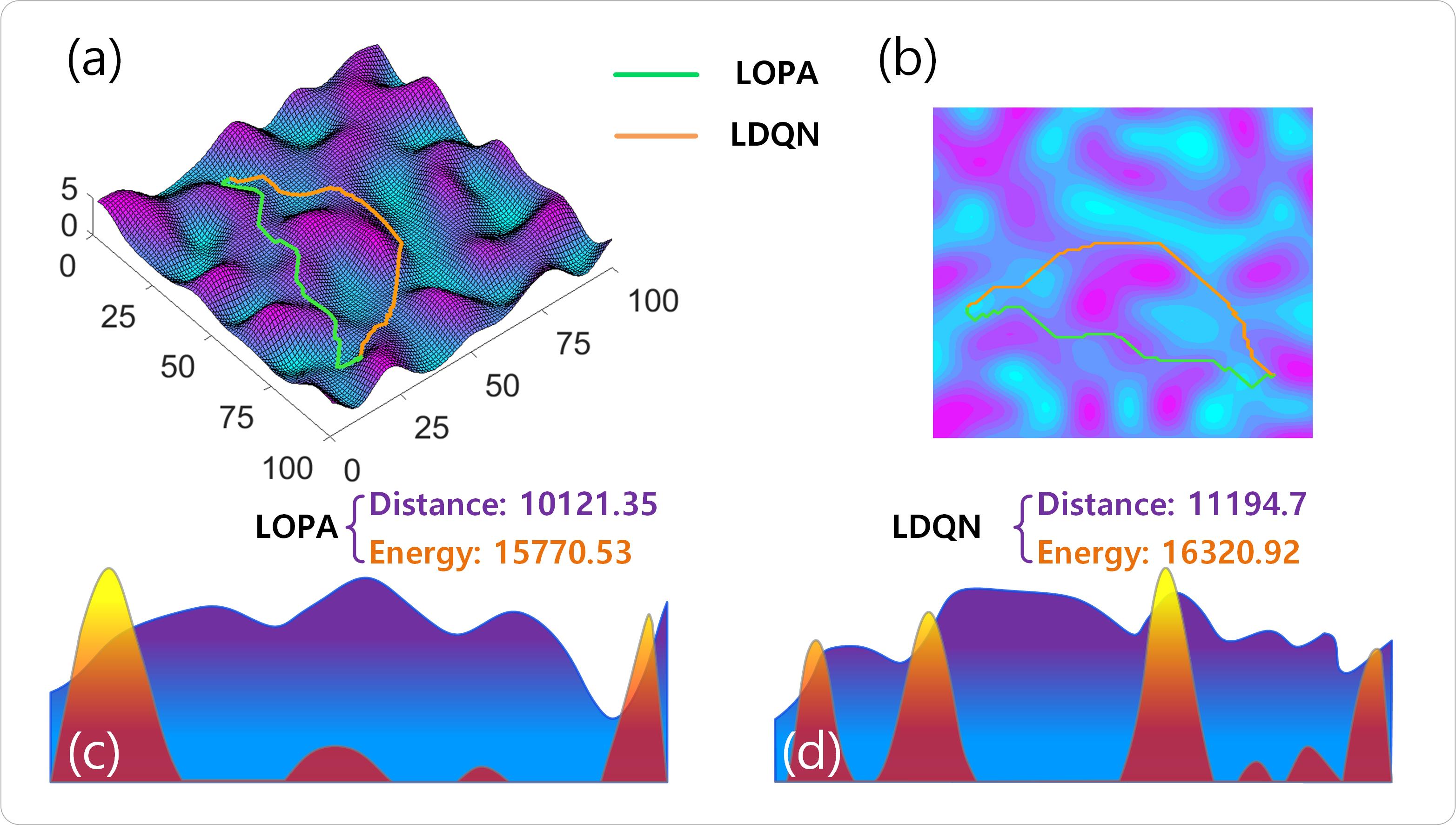}
	\caption{Case 3 of experiment 2}
\end{figure}

\begin{figure}[htp]
	\centering
	\includegraphics[width=0.98\columnwidth]{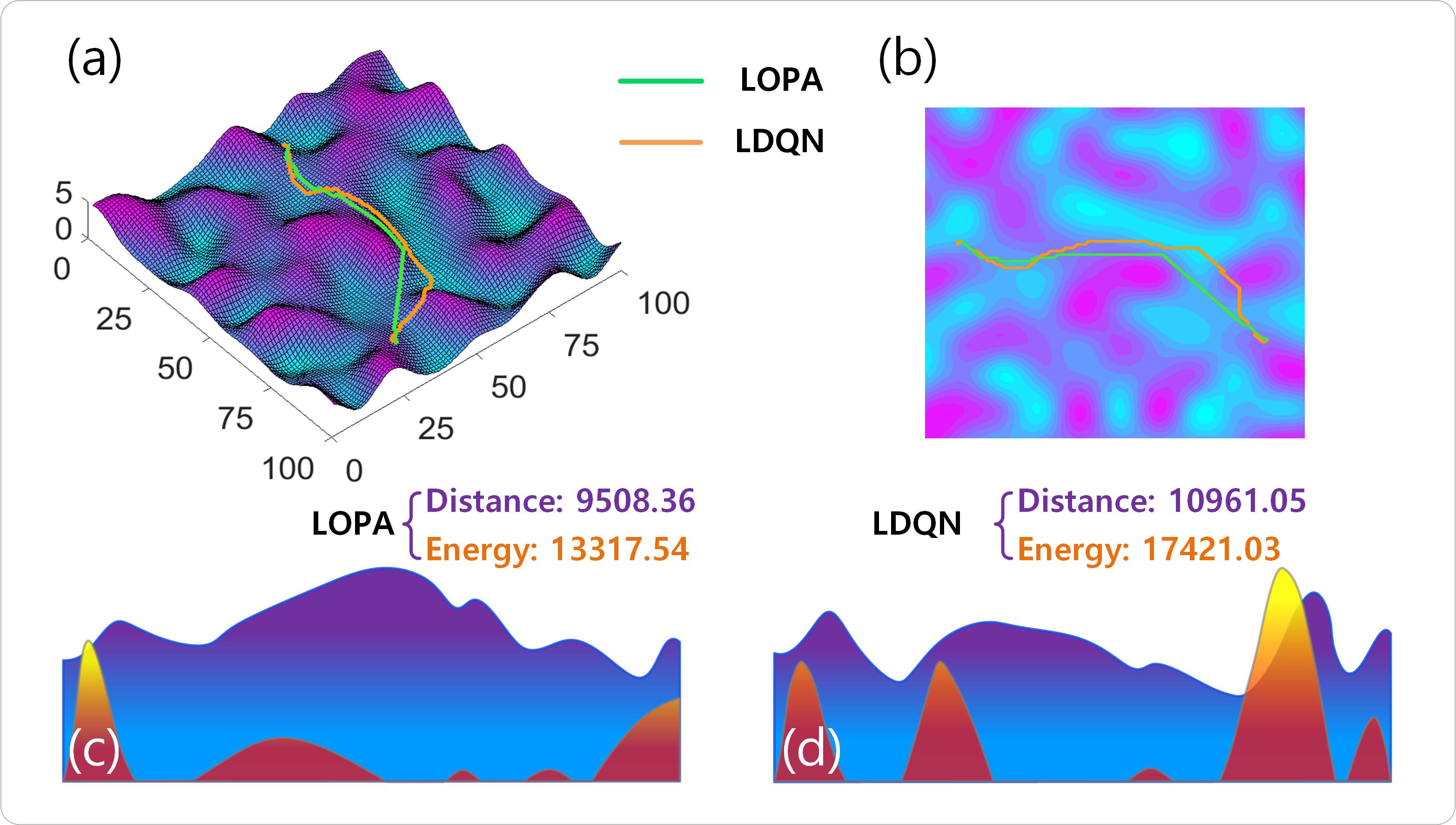}
	\caption{Case 4 of experiment 2}
\end{figure}



\subsubsection{Discussion}
Results demonstrate that the LOPA exhibit great generalization performance. This reveals that the LOPA can better understand the key terrain on the map and correctly carry out the reasoning. On the contrary, LDQN cannot use the global terrain for planning strategy reasoning and can only guarantee the reachability of the paths.

\subsection{Experiment 3}
This experiment also aims to validate LOPA's generalization performance in comparison with the traditional methods.
\subsubsection{Setting}
Planning efficiency of LOPA was evaluated by comparing its performance against an improved A* method, H3DM, and improved RRT method. The heuristic function of the A* method was specifically designed for the 2.5D planning task, utilizing 3D curvilinear distance, linear distance, and energy consumption estimation as models. We also improved the node selection strategy of the RRT method to avoid positions at higher peaks and lower valleys. A total of 50 maps with a size of 100×100 were used to train the LOPA for multitask evaluation, and four maps were selected from the test set. Results are presented in Fig. 9 through Fig. 12, in which sub-figures (a) and (b) display planned paths on 3D perspectives, while sub-figures (c) and (d) depict planned paths in 2D perspectives. Sub-figures (e) and (f) illustrate search areas for the A* and RRT methods, respectively. Quantitative results can be found in TABLE III through TABLE VI. It should be noted that LOPA utilized GPU during testing, whereas we consider the ability to utilize GPU as one of its advantages.

\begin{figure}[htp]
	\centering
	\includegraphics[width=0.95 \columnwidth]{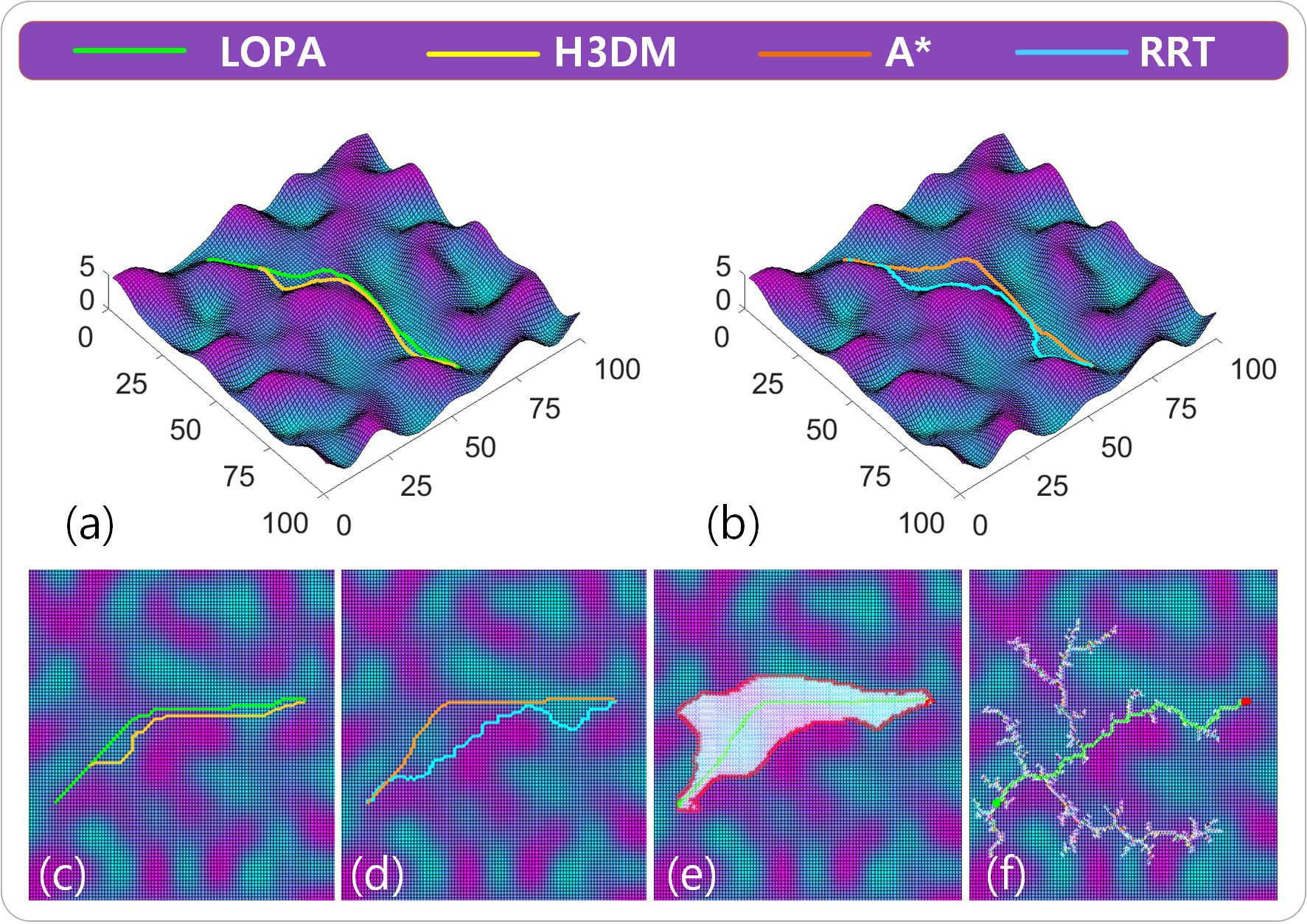}
	\caption{Case 1 of experiment 3}
\end{figure}

\begin{figure}[htp]
	\centering
	\includegraphics[width=0.95 \columnwidth]{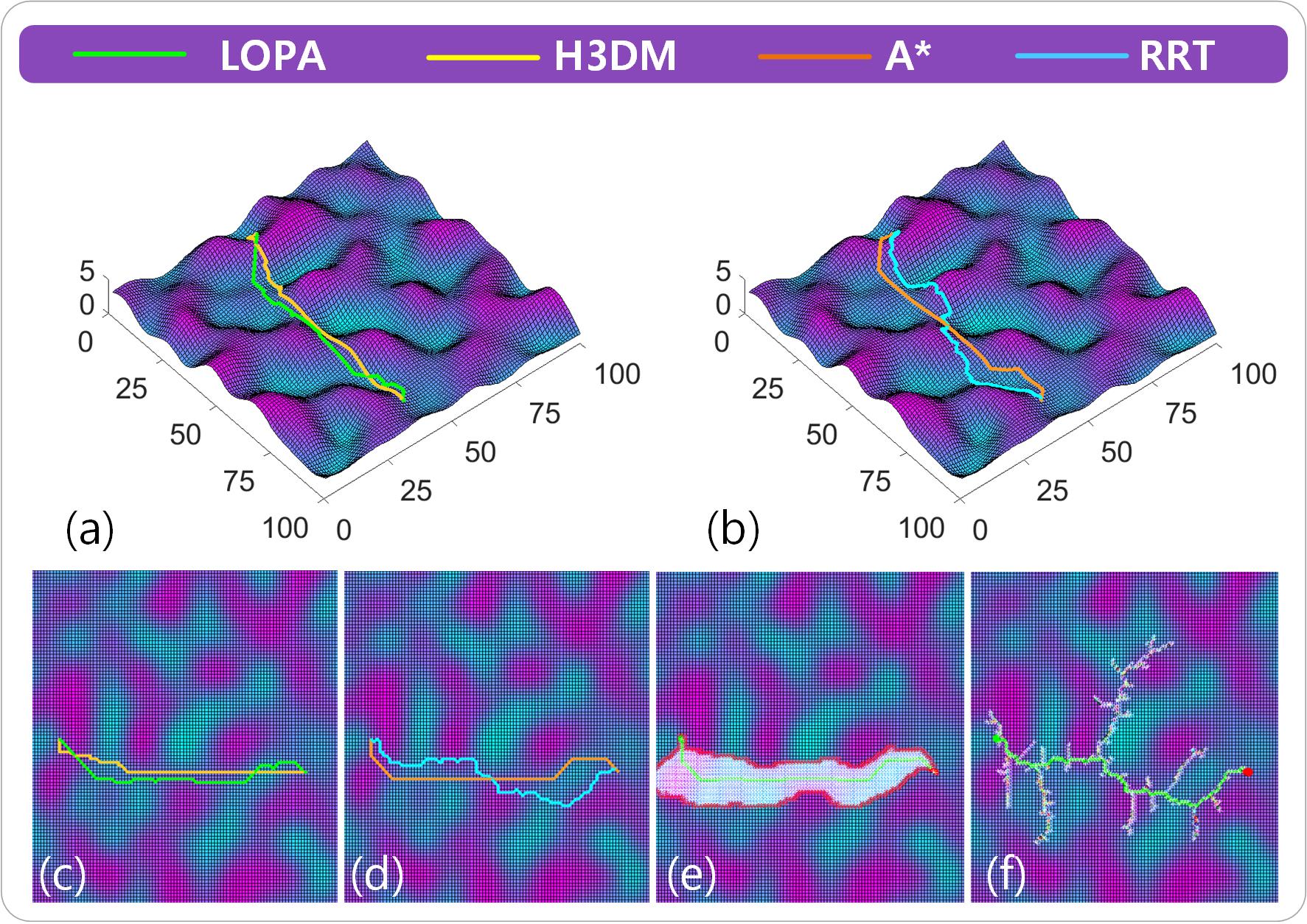}
	\caption{Case 2 of experiment 3}
\end{figure}

\begin{figure}[htp]
	\centering
	\includegraphics[width=0.95 \columnwidth]{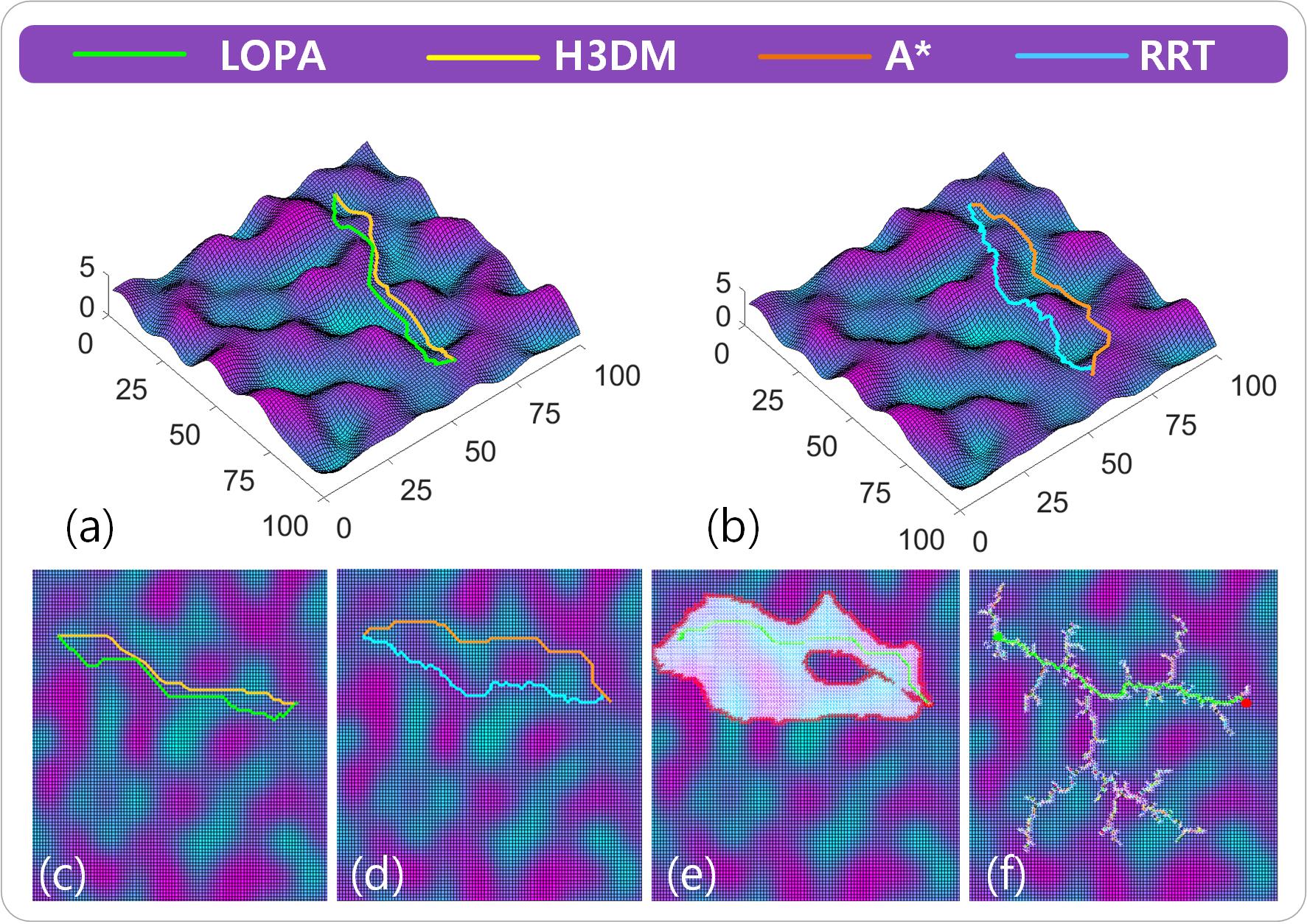}
	\caption{Case 3 of experiment 3}
\end{figure}

\begin{figure}[htp]
	\centering
	\includegraphics[width=0.95 \columnwidth]{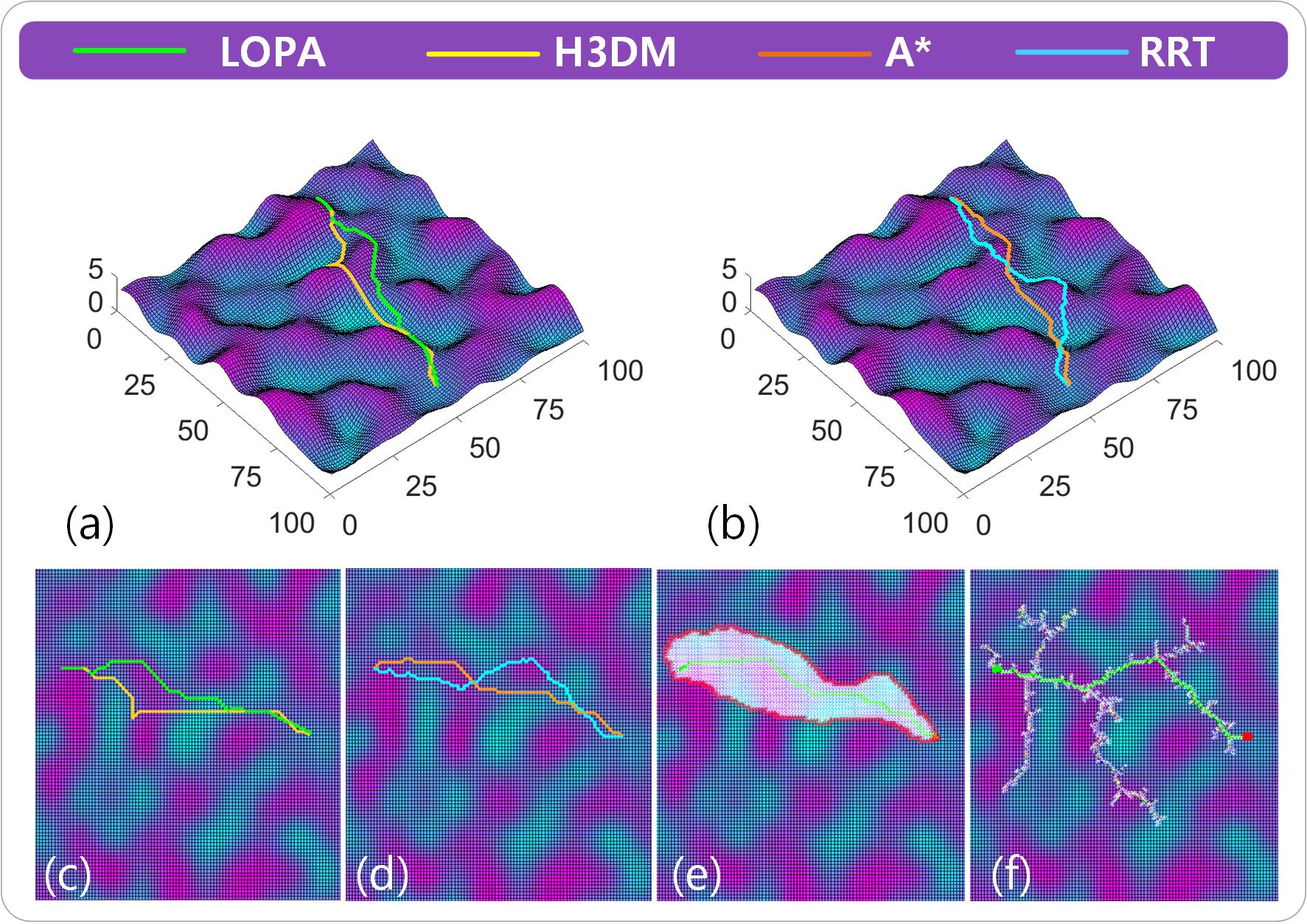}
	\caption{Case 4 of experiment 3}
\end{figure}



\begin{table}[htp]
	\fontsize{8}{8}\selectfont   
	\centering  
	\renewcommand\arraystretch{1.5}
	\caption{Quantitative result of Case 1}
	\begin{tabular}{lcccc}
		\toprule[1.2pt] 
		\textbf{Method} & \textbf{Energy (u)} & \textbf{Distance (m)} & \textbf{Sum} & \textbf{Time (s)}  \\
		\midrule
		LOPA  & 12066.97 & 9426.84	& \textbf{21493.81} &   \textbf{2.56} \\
		H3DM  & 12359.83 & 9479.56 & 21839.39 & 5.24    \\
		\(A^*\) & 11920.72 & 9654.28 & 21575 & 9.43    \\
		RRT  & 18652.36	& 11303.66	& 29956.02	& 108.82     \\
		\bottomrule[1.2pt]
	\end{tabular}
\end{table}
\begin{table}[htp]
	\fontsize{8}{8}\selectfont   
	\centering  
	\renewcommand\arraystretch{1.5}
	\caption{Quantitative result of Case 2}
	\begin{tabular}{lcccc}
		\toprule[1.2pt] 
		\textbf{Method} & \textbf{Energy (u)} & \textbf{Distance (m)} & \textbf{Sum} & \textbf{Time (s)}  \\
		\midrule
		LOPA  & 11951.69 &	9214.49 &	21166.18 &	 \textbf{2.49} \\
		H3DM  & 12790.32 & 8838.83 &	21629.15 &	5.93    \\
		\(A^*\) & 11205.92 & 9242.6 &	 \textbf{20448.52} &	6.75  \\
		RRT  & 17479.37	& 10838.61	& 28317.98	& 30.09     \\
		\bottomrule[1.2pt]
	\end{tabular}
\end{table} 
\begin{table}[htp]
	\fontsize{8}{8}\selectfont   
	\centering  
	\renewcommand\arraystretch{1.5}
	\caption{Quantitative result of Case 3}
	\begin{tabular}{lcccc}
		\toprule[1.2pt] 
		\textbf{Method} & \textbf{Energy (u)} & \textbf{Distance (m)} & \textbf{Sum} & \textbf{Time (s)}  \\
		\midrule
		LOPA  & 12877.1	& 9532.8	& 22409.9 &	 \textbf{2.56}  \\
		H3DM  & 12604.02	& 8760.07	& \textbf{21364.09}	& 5.10  \\
		\(A^*\) & 12747.12	&9566.22	& 22313.34	& 22.73  \\
		RRT  & 16402.91	& 10398.85	& 26801.76	& 72.64   \\
		\bottomrule[1.2pt]
	\end{tabular}
\end{table}
\begin{table}[htp]
	\fontsize{8}{8}\selectfont   
	\centering  
	\renewcommand\arraystretch{1.5}
	\caption{Quantitative result of Case 4}
	\begin{tabular}{lcccc}
		\toprule[1.2pt] 
		\textbf{Method} & \textbf{Energy (u)} & \textbf{Distance (m)} & \textbf{Sum} & \textbf{Time (s)}  \\
		\midrule
		LOPA  & 11546.87	& 9343.9	& 20890.77	 &	 \textbf{2.55}  \\
		H3DM  & 12810.97	& 9387.55	& 22198.52	& 5.78  \\
		\(A^*\) & 10481.44	& 9108.63	& \textbf{19590.07}	& 9.24  \\
		RRT  & 16787.93	& 10522.25	& 27310.18	& 48.15   \\
		\bottomrule[1.2pt]
	\end{tabular}
\end{table}

\subsubsection{Result}

due to the use of GPU, LOPA exhibits the fastest planning speed among the evaluated approaches. Although A* plans paths with relatively good global perspective results, it requires approximately 10 s in most cases and up to 22 s in the slowest case. Meanwhile, H3DM is also relatively fast at around 6 s, with path globality only slightly inferior to that of LOPA in certain scenarios. Conversely, RRT is significantly slower (Case 1 requiring 108 s), with poor path quality characterized by increased length and reduced smoothness. While path globality of LOPA is inferior to that of the A*, the difference is not significant. Overall, considering the planning speed, LOPA emerges as the most efficient approach in this experiment.

\subsubsection{Discussion}

While A* represents the most effective method for 2D path planning, it requires specific design of the heuristic function for 2.5D planning tasks; otherwise, it may result in extensive area searching or sub-optimal path quality. In our study, the improved A* maintains optimal performance, although its heuristic function is rather complex and requires significant time to estimate candidate point-to-target costs. Despite the improved node selection strategy, the RRT still demands substantial search area and time consumption while delivering less smooth planning outcomes. H3DM offers pure heuristic planning with fewer search points, making it more efficient than the A* and RRT. However, this advantage comes at the cost of path globality since H3DM adopts a complex model that necessitates extensive heuristic value computation.

Compared to these methods, the LOPA does not require node-by-node traversal. With fewer search nodes and faster computation speed, LOPA achieves ultra-fast planning. Furthermore, through learning global and local terrain information, LOPA acquires comprehensive planning knowledge, ultimately resulting in high-quality paths. This experiment also suggests that the LOPA attains a greater generalization capability.

In addition, we have also tried to adopt the current famous attention mechanism \cite{2017Attention} to solve this problem. However, the results we obtained were not optimistic. During the training process, the Q network still did not learn to actively focus on the main terrain, and its generalization ability was not improved. In contrast, the method we investigated can obtain expected experimental results. Besides, the proposed attention model can be understood as a data augmentation approach, which removes the noise from data.

\section{Conclusion}
This paper propose a novel DRL method, LOPA, to improve the capability of DRL for global path planning tasks. LOPA composes of an attention model and dual-channel network. The attention model effectively transforms the original observation into local and global views, thus prioritizing key terrain features for LOPA to focus on. The dual-channel network combines these two views to enhance LOPA's reasoning capabilities. Simulation results demonstrate that LOPA shows improved convergence and generalization performance as well as great planning efficiency. 

\section*{Acknowledgment}

This work is support by National Key R\&D Program of China (2022YFB4003800),  National Natural Science Foundation of China (No.62073127).



%

%
%

\bibliographystyle{IEEEtran}
\bibliography{IEEEabrv,IEEEexample}

%

\begin{IEEEbiography}[{\includegraphics[width=1in,clip,keepaspectratio]{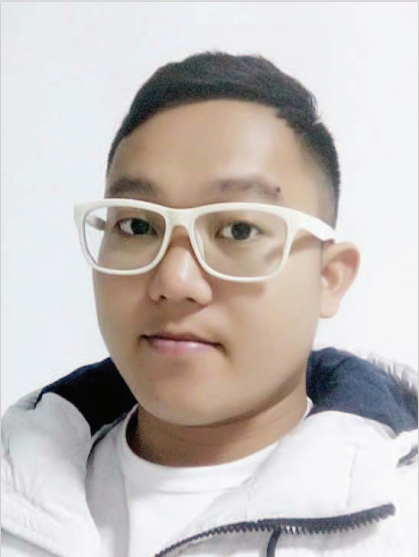}}]{Guoming Huang} received the Ph.D. degree at Hunan University in 2022. He was also a joint training of doctoral students at University of Wisconsin-Madison during 2019-2021. He is currently a researcher at Guilin University of Electronic Technology. He is passionate on research. His interests include machine learning, intelligent control, and path planning.
\end{IEEEbiography}


\begin{IEEEbiography}[{\includegraphics[width=1in,clip,keepaspectratio]{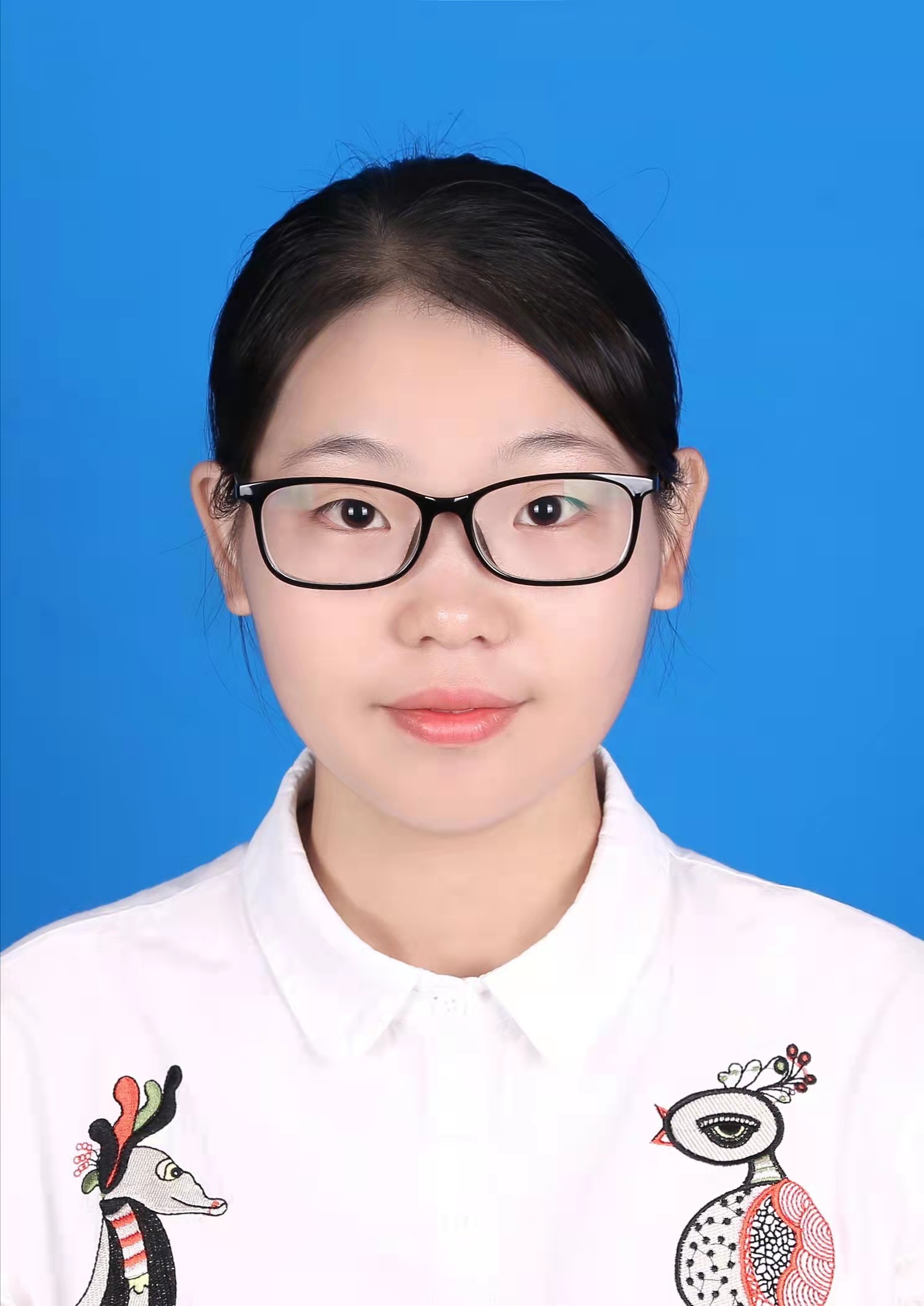}}]{Mingxin Hou} received the B.S. in Anhui Normal University in 2020. She is currently a M.S. candidate at Hunan University. Her research interests include machine learning, and intelligent path planning.
\end{IEEEbiography}


\begin{IEEEbiography}[{\includegraphics[width=1in,clip,keepaspectratio]{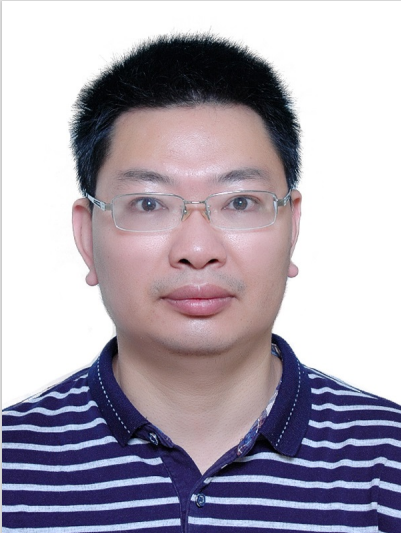}}]{Xiaofang Yuan} received the B.S., M.S. and Ph.D. degrees in electrical engineering all from Hunan University, Changsha, China, in 2001, 2006 and 2008 respectively. He is currently a professor at Hunan University. His research interests include intelligent control theory and application, industrial process control, and artificial neural networks.
\end{IEEEbiography}



\begin{IEEEbiography}[{\includegraphics[width=1in,height=1.25in,clip,keepaspectratio]{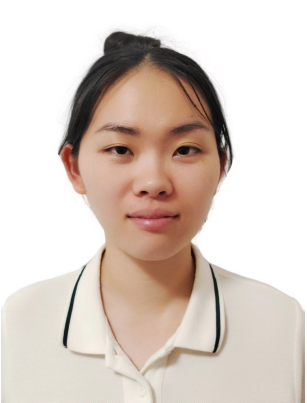}}]{Shuqiao Huang}
	is currently working toward the M.S. degree in control engineering with the School of Electronic Engineering and Automation, Guilin University of Electronic Technology. Her research interests include deep learning, and intelligent path
	planning.
\end{IEEEbiography}


\begin{IEEEbiography}[{\includegraphics[width=1in,height=1.25in,clip,keepaspectratio]{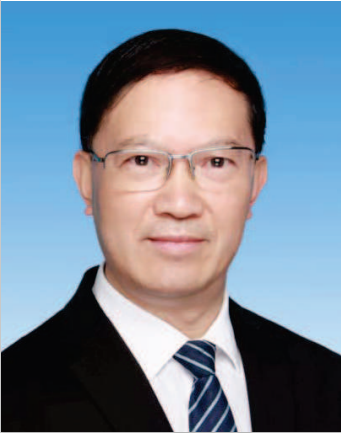}}]{Yaonan Wang}
is an academician of the Chinese Academy of Engineering. He is currently a professor and Ph.D. student supervisor at Hunan University. His research interests include intelligent robot control technology, machine vision perception and image processing technology, intelligent manufacturing equipment measurement and control technology.
\end{IEEEbiography}

\end{document}